# An End-to-End Cascaded Image Deraining and Object Detection Neural Network

Kaige Wang*[12], Tianming Wang*[12], Jianchuang Qu[2], Huatao Jiang[1], Qing Li[1], Lin Chang[1]

*Abstract*— While the deep learning-based image deraining methods have made great progress in recent years, there are two major shortcomings in their application in real-world situations. Firstly, the gap between the low-level vision task represented by rain removal and the high-level vision task represented by object detection is significant, and the low-level vision task can hardly contribute to the high-level vision task. Secondly, the quality of the deraining dataset needs to be improved. In fact, the rain lines in many baselines have a large gap with the real rain lines, and the resolution of the deraining dataset images is generally not ideally. Meanwhile, there are few common datasets for both the low-level vision task and the high-level vision task. In this paper, we explore the combination of the low-level vision task with the high-level vision task. Specifically, we propose an end-to-end object detection network for reducing the impact of rainfall, which consists of two cascaded networks, an improved image deraining network and an object detection network, respectively. We also design the components of the loss function to accommodate the characteristics of the different sub-networks. We then propose a dataset based on the KITTI dataset for rainfall removal and object detection, on which our network surpasses the state-of-the-art with a significant improvement in metrics. Besides, our proposed network is measured on driving videos collected by self-driving vehicles and shows positive results for rain removal and object detection.

## I. INTRODUCTION

Image deraining is an essential low-level computer vision task, and its performance has an obvious impact on the robustness and reliability of high-level computer vision tasks. Scenes with rain typically consist of rain lines of different lengths, widths, shapes, transparency, tilt angles, and clean backgrounds[1], sometimes with significant additional noise due to the rainfall[2]. Rainfall can have a strongly detrimental effect on various scenes; for example, rainfall is often accompanied by lower ambient brightness, where rain and fog in the distance can significantly reduce the visibility of distant buildings and rain lines in the near distance may obscure objects[3], thus severely affecting computer vision tasks. For example, raindrops streaks could cause low contrast and blurring of the visual data, which may degrade the accuracy and robustness of the object detection task and affect the functions of robot vision system. Therefore, the research and application of rain removal algorithms are of great significant for both the image processing field and robotics field.

Generally, the deraining task is divided into two categories: single-image deraining and video deraining. Single-image deraining is more challenging than video deraining because of the lack of timing information, while the application scenario of single-image deraining is wider and more universal, so this paper focuses on single-image deraining and its applications. Single image deraining has undergone the development from traditional time-domain and frequency-domain filtering to deep neural network deraining, realizing the transformation of algorithm model-driven and data-driven[4]. The deraining algorithm has been developing continuously in recent years, and the image deraining has become increasingly effective.

Object detection is one of the noticeable tasks of computer vision. It includes two sub tasks: identifying which objects and their positions in the picture, corresponding to classification and regression tasks respectively. With the development of deep learning technology, CNN has been widely used in object detection technology and achieved well results. R-CNN[5] was first used for object detection. Candidate regions were proposed through selective search. Then each candidate region was classified by CNN and the accurate location was regressed. Fast R-CNN[6] does not search in the original image, but in the feature image, which greatly speeds up the procedure. Later, the proposal of Faster R-CNN[7] further matured this technology. It proposes an RPN network, which can select the region of interest by itself. The feature map of these areas will generate category confidence and position offset through two parallel branches to form a detection box. The region proposal will overlap, and NMS is required to remove redundant detection. On the other hand, YOLO series[8][9][10] proposes a unified architecture, which can directly predict the class probability and boundary box without clearly indicating the region of interest (ROI). This method has simple structure and is very convenient to deploy. Object detection technology has been widely used in automatic driving, because only when the automatic driving vehicle can find the surrounding obstacles, can it avoid colliding with them.

However, image deraining and object detection in rainfall weather still has several problems. First, there is a huge gap between the low-level tasks of computer vision and the high-level tasks remained. Specifically, when investigating the low-level task represented by rain removal, we often focus on how to remove rain lines from images, while neglecting how raindrop removal can effectively work tightly with high-level tasks such as object detection and scene segmentation, and how to integrate with high-level tasks. Ignoring the integration with high-level computer vision tasks will undoubtedly limit the practical application of image deraining in real-world scenarios. Secondly, there is a lack of datasets for joint training of image deraining and high-level

*Kaige Wang and Tianming Wang are co-first authors.

[1]Authors are with the Institute of Microelectronics of the Chinese Academy of Sciences, Beijing, China, email: {wangkaige, wangtianming, jianghuatao, liqing, changlin}@ime.ac.cn. Corresponding author: Lin Chang.

[2]Authors are also with the University of Chinese Academy of Sciences, Beijing, China, email: qujianchuang21@mail.ucas.ac.cn.

computer vision tasks, and the existing image deraining baselines have problems such as low resolution, tiny amount of data, little variation of rainfall effect and not realistic, meanwhile, it is also hard to collect paired real with or without rain datasets in real life.

To address the limitations of previous researches, we propose an end-to-end rainfall weather object detection network that consists of a deraining sub-network and an object detection sub-network cascaded to attenuate the impact of rainfall on the object detection task. Our contributions in this paper are mainly the three points as follows.

1. We propose an end-to-end cascaded image deraining and object detection neural network, which combines several features of excellent networks in recent years, and we are the first in our knowledge to propose an end-to-end rain removal and object detection network.

2. We design a new joint network loss function, which is more effective in training than the original common loss function, and experiment results prove that it can make the deraining images closer to the rain-free images, thus improving the object detection accuracy under rainfall conditions.

3. We propose a high-resolution joint dataset for rain removal and object detection, Rain-KITTI. Adding different characteristics of rain lines to the original KITTI dataset to better fit the rainfall scenarios in autonomous driving.

## II. RELATED WORK

### A. Development of single image deraining

The rain removal task is typically divided into video rain removal and single image rain removal, and has undergone a transition from model-driven to data-driven development.

Video deraining mainly includes time-domain video deraining, Low Rank and Sparsity video deraining and deep learning video deraining approaches. The principle of time-domain video deraining method represented by [11][12] is utilizing a space-time correlation model to capture the dynamics of raindrop. The principle of low-rank sparse video deraining method represented by [13][14] is using low rank and sparsity properties, and its enables to obtain better deraining effect. In recent years, the deep learning video deraining method represented by [15][16] has been developed greatly.

Single image deraining is considered more challenging than video deraining due to the lack of temporal information. Early image deraining commonly used frequency domain deraining methods[17][18] and a priori knowledge-based deraining methods[19][20], which have achieved positive results in some scenarios. However, these methods have high computational complexity, long inference time, and poor transferability. Benefiting from the powerful fitting capabilities of deep neural networks, single image deraining based on deep learning has been developed rapidly in recent years. Deep learning-based single image deraining is frequently used to extract rain lines [21][22] from rain images by the method of encoder-decoder, and uses multiple stages to extract rain lines step by step, thereby attenuates the rain potential [23][24] in the rain image. In the encoder-decoder procedure, some of the work uses LSTM module[25], channel attention module[23], and feature fusion module[24] to enhance the performance of rain removal network.

### B. Development of object detection tasks

Object detection technology based on CNN can be divided into two categories: single-stage detector and multi-stage detector. Multi-stage detector is a mainstream paradigm and has dominated object detection technology for many years. Such algorithms select sparse foreground proposals in many candidate regions, and then refine the confidence and location information of each proposal. In recent years, multi-stage detector has developed rapidly. Faster R-CNN[7] creatively uses convolution network to generate ROI by itself, and shares convolution network with the whole object detection network. Mask-RCNN[26] further improves the performance through multi task branching. Cascade RCNN[27] continuously screens and optimizes candidate boxes through a cascade structure.

The single-stage detector does not need proposals, but directly predicts the bounding box and classification score. Overfeat[28] is the earliest single-stage detector based on CNN. Later, Yolo series[8][9][10] and SSD[29] enriched the single-stage detection algorithm. Retinanet[30] uses focal loss to balance the imbalance between positive and negative samples, which greatly improves the accuracy of single object detection algorithm. Some point-based detection methods [31][32] solve the detection problem by identifying and grouping multiple key points. FCOS[33] completes the object detection task through anchor points and point to boundary distances.

With the development of object detection technology, datasets are constantly enriched. At present, VOC[34] and COCO[35] are most familiar. In the field of automatic driving, Kitti[36] is one of the most widely used datasets at present; Cityscape[37] focuses on the understanding of urban street scenes, including 50 stereo video sequences recorded from street scenes of different cities.

### C. Architectures for Residual Block

ResNet[38] achieved impressive results on ImageNet in 2015. As a landmark network structure in the history of neural network development, the residual network structure has attracted considerable attention and investigation, and leading to the development of many improved versions, including ResNetV2[39], Wider-ResNet[40], Dilated ResNet[41], ResNeXt[42], ResNeSt[43], etc.

Further adjustments to the order of the components in ResNet make the ResNetV2 network easier to optimize Although increasing the depth of the network can improve the results, as the network continues to deepen, each additional small improvement in performance requires a great increase in network depth. Wider ResNet found that increasing the width of the network also improves the performance of the network, and that a wide network is more computationally efficient than a deep network. CNN gradually decreases image resolution when convolution operation is performed. Therefore, Dilated ResNet proposes Dilated Convolution and introduces it into ResNet to increase the resolution of the feature map and improve the performance of the network without increasing model parameters, and verifies that it is effective in both image classification and object detection, image segmentation, etc.

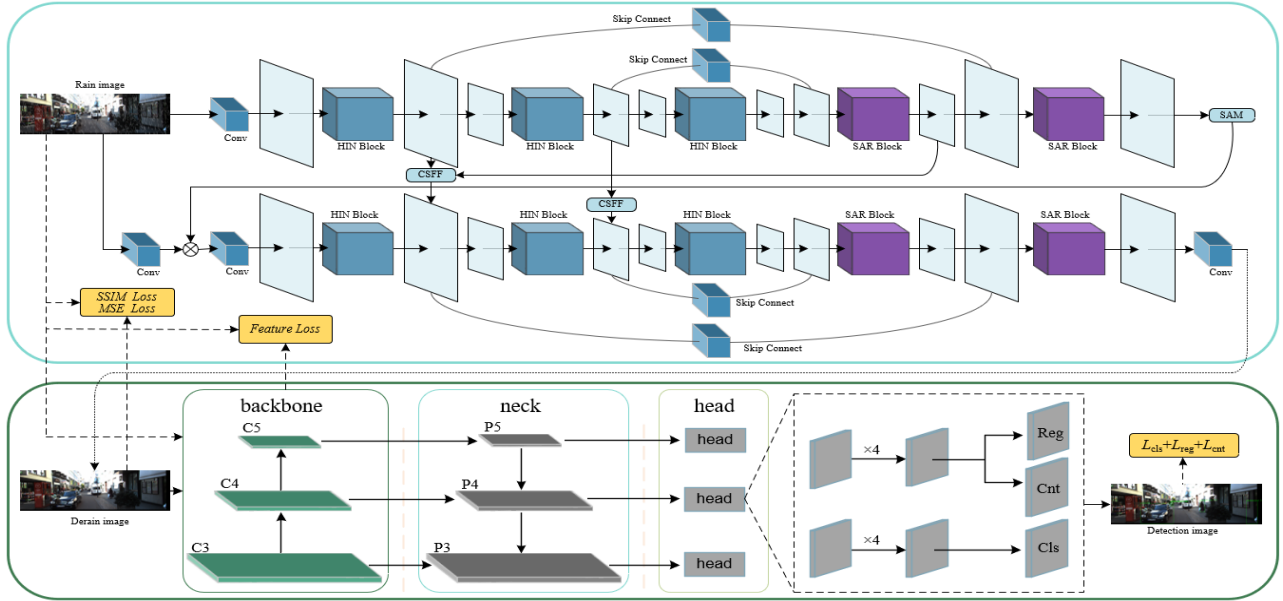

Figure 1. Our proposed rain removal and object detection model architecture

ResNeXt combines the idea of inception with ResNet and demonstrates that in addition to increasing the depth and width of the network, the performance of the model can be improved by increasing the cardinality. Further, ResNeSt proposes the Split-Attention module based on ResNeXt to further enhance the role of residual structure in model performance improvement.

III. METHODOLOGY

In this section, we introduce the proposed rain removal and object detection model by showing and analyzing the model structure, multi-stage model training strategy, and model loss function.

A. Deraining Network Architecture

The architecture of our proposed deraining network is shown in the blue cell of Fig. 1. Inspired by the MPRNet and HINet structures, our deraining network consists of two progressive stages. Each stage consists of a U-Net structure composed of encoders and decoders, where the encoder consists of HIN-blocks[24] and the decoder consists of Split-Attention Resblocks (SAR) designed by us. the down-sampling effect is implemented in the encoder and the up-sampling effect is implemented in the decoder. A skip connect is used between the decoder and the encoder to compensate for the information missing in the up-sampling process. A cross-stage feature fusion (CSFF) module and a supervised attention module (SAM)[23] are used between the different stages to connect the two stages and better fuse the cross-stage features.

Inspired by ResNeXt and ResNeSt, we use the Split-Attention method to divide the vector whose input are C channels into k Cardinal groups, each Cardinal group includes C/k channels, and each Cardinal group is subdivided into r Split groups, so there are totally k×r groups. Similar to SE module[44], our proposed structure has a Split-Attention module consisting of Global pooling layer, FC layer, and Softmax layer, etc. Split-Attention block achieves attention effects across feature maps with impressive results. In addition, we add a cross-layer connection called Res Conv from input to the output of the Split-Attention module, which allows a more efficient feature fusion and gradient flow during the model training. The Split-Attention Resblock structure we designed is shown in Fig. 2.

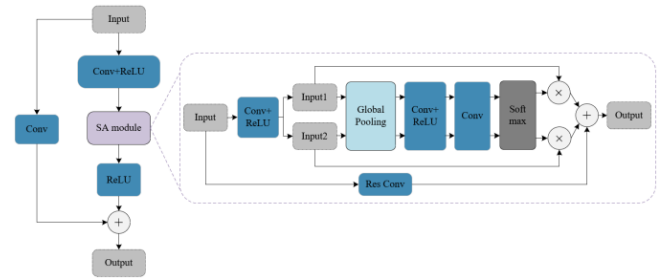

Figure 2. Our proposed Split-Attention Resblock architecture.

B. Object Detector Architecture

The detector we use is FCOS, which is a one-stage full convolution object detection algorithm based on pixel level prediction. FCOS does not rely on predefined anchor boxes or proposed areas, so it completely avoids the complex operation on the anchor box, such as calculating the overlap in the training process, and saves the memory occupation in the training process. FCOS avoids all super parameters related to anchor frame which are very sensitive to final detection results.

The structure of FCOS detector network is shown in the green cell of Fig. 1. Specifically, it uses Resnet to extract features and obtain feature maps [C3, C4, C5]. The feature map will be sent to the FPN[45] structure to obtain multi-scale feature map [P3, P4, P5, P6, P7]. After each feature-map is a detector, which contains three branches: classification, regression and center-ness. For the classification branch, a total of C predicted values are output and C binary classifications are used. For the regression branch, a 4D real vector $t = (l,t,r,b)$ is predicted for each position, and its corresponding target is the distance between the current position and the four vertices of the GT box: FCOS adds an additional center ness branch at the end of the region branch to suppress the low-quality detection frames predicted by the positions that deviate from the target center.

*C. Multi-stage Model Training Strategy*

Since the joint network includes two sub-networks, image deraining and object detection, the training parameters of the joint network will include more uncertainties if they are initialized completely randomly, which in turn makes the training converge slowly. Therefore, we train the joint network in three stages to improve the training efficiency. The detailed strategies of the three stages are as follows.

Firstly, we train the object detection network FCOS individually, the input of the network is a no-rain image, and the output of the network is compared with the class and location coordinate information of the object in the image. The reason for pre-training the object detection network in the first stage is that it can be used to train the deraining network in the second stage.

Secondly, we train the image deraining network with downstream work. The input of the network is the image with rain, and the output of the network is the deraining image and its feature map generated by the backbone network of FCOS. We expect deraining images and rain free images to maintain not only pixel consistency, but also feature consistency, which is favorable for downstream tasks.

Finally, we jointly train the network based on sub-network pre-training. The input of the network is the image with rain and the output is compared with the image without rain and the class and location information of the object in the image.

The varied training strategies in different stages will achieve less overall training epochs. After three stages of training strategy, the network thus could recover rain-free images from images with rain and to detect objects.

*D. Loss Function Design*

The total loss function of the network is calculated as shown in Eq.1. where $L_{Derain}$ denotes the loss function of the rain removal task, $L_{Downstream}$ denotes the loss function of the object detection task, $a$ denotes the weighting factor of the rain removal task, and $\beta$ denotes the weighting factor of the object detection task.

$$L_{total} = a * L_{Derain} + \beta * L_{Downstream} \quad \text{Eq.1}$$

Due to the differences in training strategies in each stage, the specific expressions of the loss function in each stage differ. Specifically, in the Stage 1, since we only train the object detection network, $a$ is set to 0 and $\beta$ is set to 1, which means that the loss function of the deraining task is not considered. $L_{Downstream}$ is expressed as the focal loss of the FCOS network, and the formula of $L_{Downstream}$ is shown in Eq.2, where $L_{cls}$ is focal loss [45] and $L_{reg}$ is GIOU loss [46]. $L_{cnt}$ denotes the BCE loss for the normalized distance from the location to the center of the object.

$$L_{Downstream} = L_{cls} + L_{reg} + L_{cnt} \quad \text{Eq.2}$$

In the Stage 2, we set $a$ to 1 and $\beta$ to 0.1 because of training the image deraining network with fused object detection features. $L_{Derain}$ uses SSIM loss, and the formula for $L_{Derain}$ is shown in Eq.3 and Eq.4, where $x$ denotes the no-rain image and $y$ denotes the deraining image, and $l(x,y)$, $c(x,y)$, $s(x,y)$ represent the lightness contrast function, contrast ratio contrast function, and structure contrast function, respectively. Inspired by the Perceptual Contrastive Loss proposed by SAPNet[49], $L_{Downstream}$ uses the Feature Maps loss (FML) $L_{feature}$, and feature maps are obtained from the backbone of the FCOS by inputting the no-rain and deraining images. $L_{feature}$ is consistent with Eq.5, where $f$ and $y$ represent the feature maps and deraining images.

$$L_{Derain} = 1 - SSIM(x,y) \quad \text{Eq.3}$$

$$SSIM(x,y) = [l(x,y)]^{\alpha}[c(x,y)]^{\beta}[s(x,y)]^{\gamma} \quad \text{Eq.4}$$

$$L_{Downstream} = L_{feature} = MSE(f,y) \quad \text{Eq.5}$$

In the Stage 3 for joint training, we set $a$ to 1 and $\beta$ to 0.5. Different from Stage 2, $L_{Derain}$ uses MSE loss and SSIM loss, and the formula is consistent with Eq.6. $L_{Downstream}$ uses the focal loss proposed in Stage 1, and the formula is consistent with Eq.2.

$$L_{Derain} = 1 - SSIM(x,y) + MSE(x,y) \quad \text{Eq.6}$$

IV. DERAINING AND OBJECT DETECTION DATASET

The existing synthetic rain removal dataset has drawbacks such as small number of images, poor rain line changes, unrealistic images with rain, etc. The widespread object detection datasets used in the autonomous driving field are KITTI datasets, etc. Therefore, to narrow the gap between the rain removal task and the object detection task, we add synthetic rain lines to the KITTI dataset, and then develop a dataset for joint training of the rain removal task and the object detection task, the Rain-KITTI dataset.

*A. Components of Synthetic Raindrop*

Aiming to describe the differences in the rainfall process as accurately as possible, we divide the raindrops into three categories with different lengths: long, medium, and short raindrops. Each category of raindrops can be adjusted by Python code for its distribution density, raindrop angle, and raindrop width, so as to more closely approximate the real rainfall environment.

*B. Details of Rain-KITTI*

The KITTI dataset includes 7481 irrelevant images with inconsistent resolution. To adapt to the requirements of the object detection network, we padding the KITTI dataset images to a uniform resolution of 1280*384 and add raindrops. Meanwhile, the method of cutting images into patches commonly used in deraining datasets is not applicable to object detection task, so we choose full-size images without cutting as the dataset.

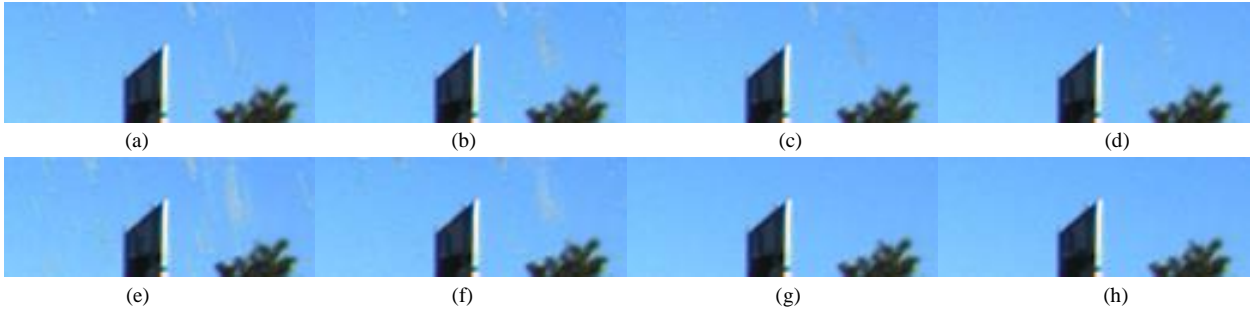

(a) (b) (c) (d)

(e) (f) (g) (h)

Figure 3. Comparison of rain removal results of different networks. (a), (b), (c), (d), (e), (f) and (g) illustrate the deraining results of DCSFnet, HINet, MARDNet, MPRNet, SAPNet, PReNet and our proposed network separately, and (h) illustrates the no-rain image.

## V. EXPERIMENTS

We tested the deraining performance of MPRNet[23], HINet[24], DCSFnet[47], MARDNet[48], SAPNet[49], PReNet[50] and our proposed network on our proposed dataset, and tested the object detection performance of the above networks. The evaluation metrics of network performance include Peak Signal to Noise Ratio (PSNR), Structural Similarity (SSIM), mean Average Precision (mAP) and mean Average Recall (mAR).

### A. Implementation Details

The network is trained on NVIDIA Tesla V100 GPU with Adam optimizer. The learning rate decay strategy is to halve the learning rate at the 30th epoch of pre-training, and no learning rate decay strategy is set for joint training. 50 epochs are trained in pre-training and 20 epochs are trained in joint training. The input of our proposed network is a full-size image during both pre-training and joint training. Consequently, the full-size image input to the network causes a significant increase in memory usage, so the batch size is set to 4 to prevent memory overflow. We used 5000 pairs of rain and no-rain images for the training dataset and 1400 pairs of rain and no-rain images for the test dataset.

### B. Metrics

PSNR is commonly used to evaluate the quality of the processed image compared with the standard image, and its algorithm principle is based on the difference between the corresponding pixel values. SSIM indicates the structural similarity, which compares the structure of the deraining image and the reference image. Calculating SSIM is mainly based on three comparisons: brightness, contrast comparison and structure.

mAP is a commonly used evaluation metric in the field of object detection. This value is related to the precision and recall of the detection object. AR (Average Recall) refers to the maximum recall in a given fixed number of test results in each picture and mAR is the mean value of AR in all categories

### C. Comparison Experiments Results

We show the effectiveness of our proposed network on the Rain-KITTI dataset in Table I, Fig. 3 and Fig. 4. Table I shows that our proposed network performs outstandingly among all networks, significantly improving the performance of the technique. Compared to the recent best algorithm HINet, we obtained a performance gain of 1.69 dB and the SSIM is closer to the ideal value 1. Compared to baseline MPRNet, which has been outstanding and commonly used in recent years, we obtained a performance gain of 2.29 dB and a significant improvement in the SSIM metric. Meanwhile, the mAP and mAR metrics improve significantly, proving that our joint network and training strategy are effective. Fig. 3 shows the comparison of rain removal results of different networks. The visual performance shows that our proposed network has the most effective removal of raindrops and approaches the original no-rain image most closely.

TABLE I: COMPARISON OF RAIN REMOVAL AND OBJECT DETECTION METRICS OF DIFFERENT NETWORKS

| Model | Deraining and Detection Evaluation Metrics | | | |
|---|---|---|---|---|
| | PSNR | SSIM | mAP | mAR |
| DCSFnet | 41.34 | 0.9895 | 0.5463 | 0.6485 |
| HINet | 41.02 | 0.9909 | 0.5463 | 0.6489 |
| MARDNet | 39.14 | 0.9831 | 0.5416 | 0.6463 |
| MPRNet | 40.42 | 0.9879 | 0.5445 | 0.6470 |
| SAPNet | 36.66 | 0.9715 | 0.5407 | 0.6451 |
| PReNet | 38.36 | 0.9874 | 0.5442 | 0.6487 |
| **Ours** | **42.71** | **0.9940** | **0.5709** | **0.6606** |

Fig. 4 shows that our proposed network achieves progress in single image deraining visualization. Compared with other networks, it can effectively remove rain patterns of different directions, sizes, and widths, and the removal of raindrops is more effective. Besides, the detailed parts of the image background are kept more clearly.

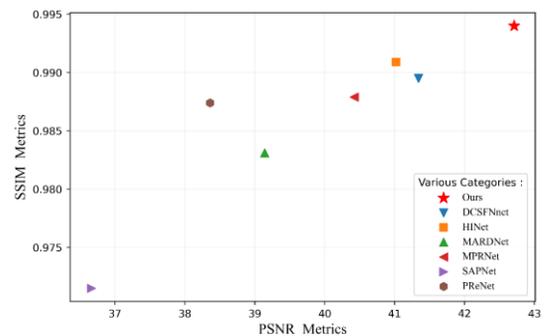

Figure 4. Comparison of rain removal metrics of different networks.

We present the efficiency of the joint network on the object detection task in Table I and Fig. 5. Table 1 shows the object detection results of different deraining networks with deraining images fed into the FCOS object detection network.

Compared with the above results, our joint deraining network trained with a multi-stage training strategy performs better. Specifically, we show in Fig. 5 the visual results of deraining and detection work. It can be found that compared with rainy images, our network can effectively attenuate the visual effect of raindrops and can better identify vehicles with higher accuracy. It is worth to be noticed that our end-to-end network also has better recognition results for distant blurred vehicles and partially obscured vehicles.

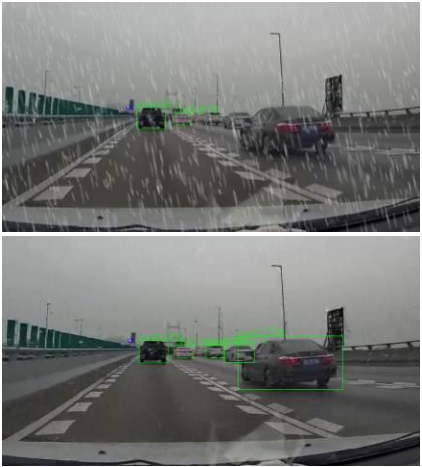

Figure 5.    The result of rain removal and object detection in a real scenario by our joint end-to-end network. Please note that besides artificially added raindrops, the image itself contains natural raindrops.

*D. Ablation Study Results*

In this section, we design ablation experiments to demonstrate the effectiveness of the innovation ideas proposed in this paper. Specifically, our proposed network is improved from HINet, so we trained and tested the performance metrics of both networks on the Rain-KITTI dataset separately to demonstrate the effectiveness of our proposed Split-Attention Resblock.

Our novelties include designing loss functions to improve the rain removal performance of the joint network. Therefore, we trained with three different loss functions on the Rain-KITTI dataset and tested their effects on the rain removal metrics of the joint network separately to demonstrate that our proposed loss functions are more effective on the rain removal performance of the network.

Metric results of image deraining ablation experiments for different structures(S) are shown in Table II, where HINet indicates our baseline, SAR indicates the Split-Attention Resblock, FML indicates Feature Map loss, and MSE indicates Mean Square Error loss.

TABLE II:   METRIC RESULTS OF IMAGE DERAINING ABLATION EXPERIMENTS FOR DIFFERENT STRUCTURES(S)

|       | S1    | S2    | S3    | Ours   |
|-------|-------|-------|-------|--------|
| HINet | ✓     | ✓     | ✓     | ✓      |
| SAR   |       | ✓     | ✓     | ✓      |
| FML   |       |       | ✓     | ✓      |
| MSE   |       |       |       | ✓      |
| PSNR  | 41.02 | 41.12 | 41.17 | **42.71** |
| SSIM  | 0.9909| 0.9926| 0.9927| **0.9940** |

## VI. CONCLUSION

In this work, we propose an end-to-end cascaded image deraining and object detection neural network. Specifically, we replace the traditional Resnet Block in U-net with a modified Split-Attention Resblock with excellent results and design a new joint loss function to improve the performance metrics of rain removal and object detection. Meanwhile, we propose a joint dataset for rain removal and object detection tasks to bridge the gap between the low-level and high-level tasks in computer vision. The rain removal performance of our network is superior in our proposed joint dataset, which also leads to improved object detection performance in rainfall environment for autonomous vehicles.

ACKNOWLEDGMENT

This research was supported by Information Center, Institute of Microelectronics of the Chinese Academy of Sciences.